# Credit Card Fraud Detection Using Autoencoder Neural Network

Ping Jiang (M.Eng)

*Department of Electrical & Computer Engineer*

*University of Western Ontario*

pjiang28@uwo.ca

*250985261*

Jinliang Zhang (M.Eng)

*Department of Electrical & Computer Engineer*

*University of Western Ontario*

jzhan964@uwo.ca

*250919668*

Junyi Zou (M.Eng)

*Department of Electrical & Computer Engineer*

*University of Western Ontario*

jzou44@uwo.ca

*250833154*

*Abstract*—**Imbalanced data classification problem has always been a popular topic in the field of machine learning research. In order to balance the samples between majority and minority class. Oversampling algorithm is used to synthesize new minority class samples, but it could bring in noise. Pointing to the noise problems, this paper proposed a denoising autoencoder neural network (DAE) algorithm which can not only oversample minority class sample through misclassification cost, but it can denoise and classify the sampled dataset. Through experiments, compared with the denoising autoencoder neural network (DAE) with oversampling process and traditional fully connected neural networks, the results showed the proposed algorithm improves the classification accuracy of minority class of imbalanced datasets.**

## I. INTRODUCTION

Credit card fraud is a growing threat with far reaching consequences in the finance industry, corporations and government. Fraud can be defined as criminal deception with intent of acquiring financial gain. As credit card became the most popular method of payment for both online and offline transaction, the fraud rate also accelerates. The main reasons for fraud is due to the lack of security, which involves the use of stolen credit card to get cash from bank through legitimate access. This results in high difficulty of preventing credit card fraud.

So how to do fraud detection is very significant. A lot of researches have been proposed to the detection of such credit card fraud, which account for majority of credit card frauds. Detecting using traditional method is infeasible because of the big data. However, financial institutions have focused their attention to recent computational methodologies to handle credit card fraud problem.

Classification problem is one of the key research topics in the field of machine learning. Currently available classification methods can only achieve preferable performance on balanced datasets. However, there are a large number of imbalanced datasets in practical application. For the fraud problem, the minority class, which is the abnormal transaction, is more important [1]. For instance, when minority class accounts for less than 1 percent of the total dataset, the overall accuracy reaches more than 99% even though all the minority class has been misclassified.

Minority class sampling is a common method to handle with the imbalanced data classification problem. The main purpose of oversampling is to increase the number of minority class samples so that the original classification information can get better retention. Therefore, in the fields where there is higher demand for the classification accuracy, oversampling algorithm is chosen in general.

This paper seeks to implement credit card fraud detection using denoising autoencoder and oversampling. For imbalanced data, we decided use above method to achieve proper model.

## II. RELATED WORKS

Data mining technique is one notable methods used in solving fraud detection problem. This is the process of identifying those transactions that are belong to frauds or not, which is based on the behaviors and habits of cardholder, many techniques have been applied to this area, artificial neural network [2], genetic algorithm, support vector machine, frequent item set mining, decision tree, migrating birds optimization algorithm, Naïve Bayes. A comparative analysis of logistic regression and Naïve Bayes is carried out in [3]. The performance of Bayesian and neural network [4] is evaluated on credit card fraud data. Decision tree, neural networks and logistic regression are tested for their applicability in fraud detections [5].

In a seminar work, [6] proposes two advanced data mining approaches, support vector machines and random forests, together with logistic regression, as part of an attempt to better detect credit card fraud while neural network and logistic regression is applied on credit card fraud detection problem [7]. A number of challenges are associated with credit card detection, namely fraudulent behavior profile is dynamic, that is fraudulent transactions tend to look like legitimate ones; credit card transaction datasets are rarely available and highly imbalanced (or skewed); optimal feature (variables) selection for the models; suitable metric to evaluate performance of techniques on skewed credit card fraud data. Credit card fraud detection performance is greatly affected by type of sampling approach used, selection of variables and detection technique(s) used.

## III. BACKGROUND

### 3.1 Autoencoder

#### A. Traditional Autoencoder Neural Network (AE)

Autoencoder is an artificial neural network used for unsupervised learning. The aim of autoencoder is to learn representations to reconstructs features for a set of data, typically for the purpose of dimensionality reduction. The simplest form of an autoencoder is a feedforward, non-recurrent neural network which is similar to the multilayer perceptron [8]. As the figure 1 shown, it has 2 parts: one is encoder and the other is decoder which are consist of by an input layer, one or more hidden layers and an output layer. The significant difference between autoencoder and multiplayer perceptron is that the output layer of autoencoder has the same number of neurons as the input layer. The purpose is to reconstruct its own inputs instead of predicting the target value from the given inputs.

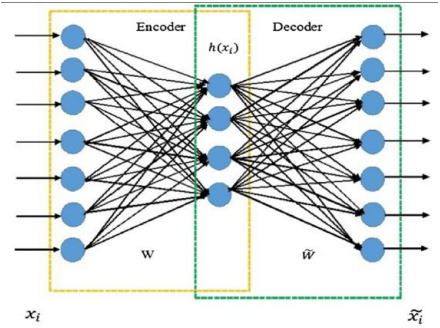

Fig. 1 architecture of autoencoder neural network

In autoencoder, the network structure has connections between layers, but has no connection inside each layer, $x_i$ is input sample, $\hat{x}_i$ is output feature.

The training of autoencoder neural network is to optimize reconstruction error using the given samples. The cost function of autoencoder neural network defined in the project is (1)

$$J_{A,E} = \frac{1}{m} \sum_{i=1}^{m} \left( \frac{1}{2} \ \|\hat{x}_i - x_i\|^2 \right) \quad (1)$$

where m represents number of input samples.

#### B. Denoising Autoencoder Neural Network (DAE)

For human, when people see an object, if there is a small part of the object is blocked, they can still recognize it. But how the autoencoder does for the "contaminated" data? There is a variation of traditional autoencoder named denoising autoencoder which could make autoencoder neural network learn how to remove the noise and reconstruct undisturbed input as much as possible [9].

As shown in figure 2, the original data is x, and $\tilde{x}$ is the data corrupted with noise. Through the complete process of denoising autoencoder, the output is $\hat{x}$. The loss function tries to minimize the difference between the output and the original data so that the autoencoder has the ability of eliminating the influence of noise and extracting features from the corrupted data. Therefore, the features generated from the learning of input corrupted with noise are more robust, which improved the data generalization ability of autoencoder neural network model to input data.

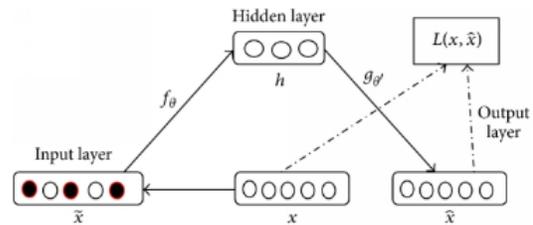

Fig. 2 Denoising autoencoder neural network

The commonly used noises are Gaussian noise, and Salt and pepper noise. And the cost function of denoising autoencoder neural network is defined according to (2)

$$J_{DA,E} = \frac{1}{m}\sum_{i=1}^{m}\left(\frac{1}{2}\ \|\hat{x}_i - x_i\|^2\right) \quad (2)$$

where $\hat{x} = f(\sum(w\tilde{x} + b))$, w represents weights and b represents bias.

*3.2 Oversampling*

Imbalanced dataset is a common problem faced in machine learning, since most traditional machine learning classification model can't handle imbalanced dataset. High misclassification cost often happened on minority class, because classification model will try to classify all the data sample to the majority class.

Oversampling is a technique used to deal with imbalanced dataset, its subject to create specific class sample so the class distribution of the original dataset can be balanced. The benefit of using oversampling is shown in figure 3.

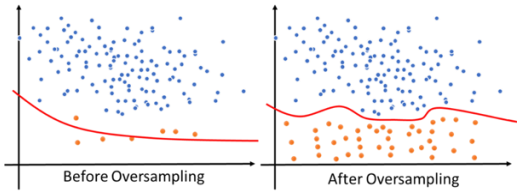

Fig. 3 Benefit of using oversampling

SMOTE (Synthetic Minority Oversampling Technique) is one of the most popular oversampling technique. In order to create a synthetic data point, first we need to find a k-nearest-neighbors cluster in the feature space, then randomly find a point within this cluster, finally using weighted average to "forge" the new data point.

*3.3 Classification fully connected model*

Deep fully connected neural network is often used in classification problem, with SoftMax cross entropy as the loss function, deep learning classification model can achieve very high accuracy.

The SoftMax function is often used in the final layer of a neural network-based classifier, it first calculates the exponential value of each output, then normalize all the output and let the sum of the output equal to 1. SoftMax function is often used for probability distribution transformation, since the output of SoftMax function is within range 0 to 1 that add up to 1, shown in the formula 3,

$$P(y_i|x_i;W) = \frac{e^{f_{y_i}}}{\sum_j e^{f_j}} \quad (3)$$

Entropy is a measure for information contents and could be defined as the unpredictability of an event. So, the greater the probability is, the smaller the unpredictability is, which means the information contents is also very small. If an event occurs inevitably with the probability of 100%, then the unpredictability and information content are 0. cross-entropy loss function takes advantages of feature of entropy equation, cross-entropy loss function can measure the goodness of a classification model, which is shown in formula 4,

$$J(\theta) = -\frac{1}{m}\sum_{i=1}^{m}\sum_{j=1}^{k}1\{y_i = j\}log\frac{e^{\theta_j^T x_i}}{\sum_{i=1}^{k}e^{\theta_j^T x_i}} \quad (4)$$

Cross-entropy can be used in multi-classification problems with the combination of SoftMax (do not consider regularization). Compared with quadratic loss function, cross-entropy loss function gives better training performance on neural networks.

*3.4 Model evaluation metric*

Accuracy is not sufficient to evaluate a classification model, especially for imbalanced dataset. For example, an imbalanced dataset with 99.9% of normal data and 0.1% of abnormal data, if the classification labels all the sample as normal class, the model can still achieve 99.9% accuracy. However, for anomaly detection, the detection rate of anomaly class is very important. Confusion matrix is often used in this situation.

Table 1. Confusion matrix for two-class problem

| Classification | Actual Positive Sample | Actual Negative Sample |
|---|---|---|
| predict as positive | TP | FP |
| predict as negative | FN | TN |

Recall (Detection rate) is the ratio between the number of correctly detected anomalies and the total number of anomalies, it evaluates how much of the anomalies can be detected in this classification model.

IV. METHODOLOGY

The credit card fraud transaction dataset we are using is downloaded from Kaggle, with totally 28315 transaction detail and 0.5% of them are labeled as fraud, the dataset is shown in the fig 4. The subject is to build a classification model for anomaly detection. Dataset contains only numerical input after doing PCA transformation. Features V1, V2, ... V28 are the principal components, the only features which have not been transformed with PCA are 'Time' and 'Amount'. Feature 'Class' is the response variable and it takes value 1 in case of fraud and 0 otherwise.

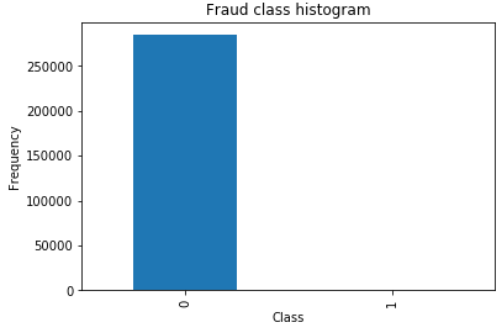

Fig. 4 Relationship between two classes

The idea is very straight forward. First, use oversampling to transform imbalanced dataset to balanced dataset. Then use denoised autoencoder to get denoised dataset. Finally using deep fully connected neural network model for final classification.

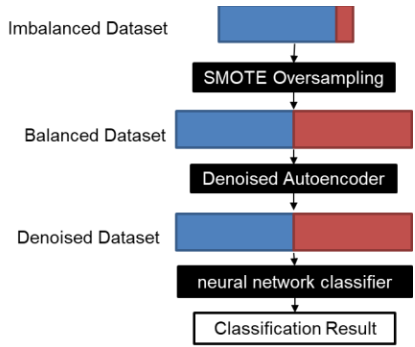

Fig. 5 Flowchart of the porcess

*4.1 Data Preprocessing*

For dataset preprocessing, drop "TIME" data, and normalized the "AMOUNT" part. Other features are obtained by PCA, do not need to do normalization. Then choose the test sample, which account for 20% of the total sample.

*4.2 Oversampling*

Our group only perform oversampling on the training dataset. Before oversampling, there are total 22652 transaction records in training dataset, with 22538 samples in normal class and 114 samples in abnormal class. After oversampling, the training dataset contains 22538 samples in normal class and 22538 samples in abnormal class.

*4.3 Denoising autoencoder*

Our group designed a 7 layers autoencoder for dataset denoising process. After we got balanced training dataset from oversampling, we add Gaussian noise to the training dataset, then feed the training dataset into this denoised autoencoder. After training this denoised autoencoder model, this autoencoder has the capability to denoise the testing dataset in the prediction process.

Table 2. Model design for denoised autoencoder

| Dataset with noise (29) |
| --- |
| Fully-Connected-Layer (22) |
| Fully-Connected-Layer (15) |
| Fully-Connected-Layer (10) |
| Fully-Connected-Layer (15) |
| Fully-Connected-Layer (22) |
| Fully-Connected-Layer (29) |
| Square Loss Function |

*4.4 Classifier*

Our group designed a 6 layers autoencoder for dataset denoise process. After we got denoised training dataset from denoised autoencoder, we feed the training dataset into this deep fully connected neural network classifier. In the end, we are using SoftMax with cross-entropy as the loss function for final classification.

Table 3. Model design for classifier

| Denoised Dataset (29) |
| --- |
| Fully-Connected-Layer (22) |
| Fully-Connected-Layer (15) |
| Fully-Connected-Layer (10) |
| Fully-Connected-Layer (5) |
| Fully-Connected-Layer (2) |
| SoftMax Cross Entropy Loss Function |

## V. EVALUATION AND RESULTS

This section first discusses the implementation details, then presents evaluation results comparing the oversampling model with model without oversampling.

*5.1 Implementation details*

Our group using built-in function from "sklearn" package for dataset normalization, and built-in function "SMOTE" from "imblearn" package for oversampling. In addition, we implement the denoised autoencoder model and deep fully connected neural network classifier with "TensorFlow". We choose "TensorFlow" because its capable of GPU acceleration. All models are trained on GTX 1060 discrete GPU w/6GB GDDR5 graphics memory. It took 10 minutes for each model to converge.

## 5.2 Results

After the training process, we perform evaluation process using another separated evaluation dataset. the accuracy rate and recall rate are applied to evaluate the accuracy of each model. The results are shown in the fig 6 and fig 7.

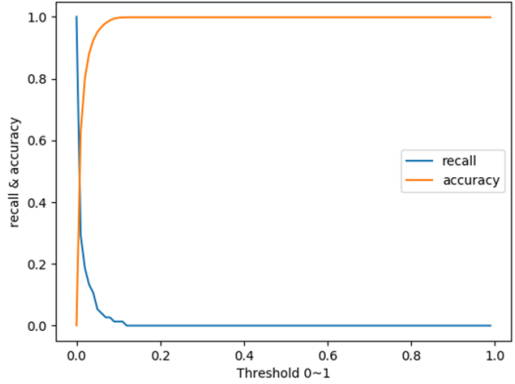

Fig. 6 Result for model 1

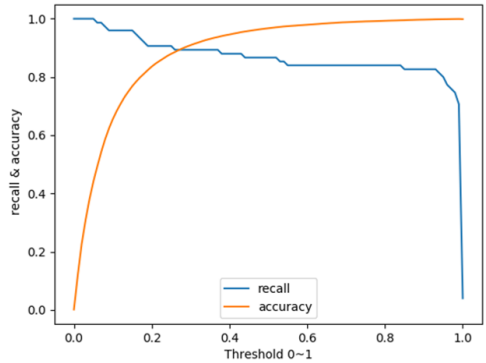

Fig. 7 Result for model 2

For model 1 without the usage of oversampling and autoencoder, the recall rate is very low, because the model classifies all the sample as normal, which means most fraud transaction is not detected. For model 2 with oversampling and autoencoder, the recall rate is acceptable, which means most fraud transaction can be detected. Some evaluation result of model 2 is showed in Table 4.

Table 4. Model 2 Evaluation Result

| Threshold | Recall Rate | Accuracy |
| --- | --- | --- |
| 0.2 | 90.66% | 83.56% |
| 0.3 | 89.33% | 90.93% |
| 0.4 | 88% | 94.58% |
| 0.5 | 86.66% | 96.73% |
| 0.6 | 84% | 97.93% |

## VI. CONCLUSION

In machine learning area, imbalance data classification receives increasing attention as big data become popular. On account of the drawbacks of traditional method, oversampling algorithm and autoencoder can be used. This study combined stacked denoising autoencoder neural network with oversampling to build the model, which can achieve minority class sampling on the basis of misclassification cost, and denoise and classify the sampled datasets. The proposed algorithm increases classification accuracy of minority class compared to the former methods, we can achieve different accuracy by controlling the threshold. In this study, when threshold equal to 0.6, we can achieve the best performance, which is 97.93%. However, the dimensionality reduction of high-dimensional data still need to be further researched.